\documentclass[letterpaper, 10 pt, conference]{ieeeconf}  

\usepackage{booktabs}
\usepackage{tabularx} 
\usepackage{graphicx}
\usepackage{url}
\usepackage{balance}

\IEEEoverridecommandlockouts                              

\title{\LARGE \bf
CognitiveDrone: A VLA Model and Evaluation Benchmark for Real-Time Cognitive Task Solving and Reasoning in UAVs
}

\author{Artem Lykov$^{*}$,  Valerii Serpiva$^{*}$, Muhammad Haris Khan, Oleg Sautenkov,\\ Artyom Myshlyaev, Grik Tadevosyan, Yasheerah Yaqoot, and Dzmitry Tsetserukou
\thanks{$^{*}$ These authors contributed equally to this work.}
\thanks{The authors are with the Intelligent Space Robotics Laboratory, Center for Digital Engineering, Skolkovo Institute of Science and Technology. 
 {\tt\small\{Artem.Lykov, Valerii.Serpiva, haris.khan, Artyom.Myshlyaev,  oleg.sautenkov, grik.tadevosyan, yasheerah.yaqoot, d.tsetserukou\}@skoltech.ru}
}}

\begin{document}

\maketitle
\thispagestyle{empty}
\pagestyle{empty}


\begin{abstract}
This paper introduces \emph{CognitiveDrone}, a novel Vision-Language-Action (VLA) model tailored for complex Unmanned Aerial Vehicles (UAVs) tasks that demand advanced cognitive abilities. Trained on a dataset comprising over 8,000 simulated flight trajectories across three key categories—Human Recognition, Symbol Understanding, and Reasoning—the model generates real-time 4D action commands based on first-person visual inputs and textual instructions. To further enhance performance in intricate scenarios, we propose \emph{CognitiveDrone-R1}, which integrates an additional Vision-Language Model (VLM) reasoning module to simplify task directives prior to high-frequency control. Experimental evaluations using our open-source benchmark, \emph{CognitiveDroneBench}, reveal that while a racing-oriented model (RaceVLA) achieves an overall success rate of 31.3\%, the base CognitiveDrone model reaches 59.6\%, and CognitiveDrone-R1 attains a success rate of 77.2\%. These results demonstrate improvements of up to 30\% in critical cognitive tasks, underscoring the effectiveness of incorporating advanced reasoning capabilities into UAV control systems. Our contributions include the development of a state-of-the-art VLA model for UAV control and the introduction of the first dedicated benchmark for assessing cognitive tasks in drone operations. The complete repository is available at \url{https://cognitivedrone.github.io}.
\end{abstract}

\section{Introduction}

In an era marked by rapid advancements in robotics and artificial intelligence, enabling robots to perform a wide range of complex tasks in dynamically changing environments has emerged as a critical challenge. Cognitive robotics strives not only to endow machines with precise control but also to equip them with high-level reasoning and decision-making capabilities that allow them to adapt to unpredictable real-world scenarios. Despite notable progress in various domains of robotics, one persistent challenge remains: how to objectively evaluate and compare cognitive robotic systems, particularly when they are expected to tackle a multitude of intricate tasks.

The scarcity of standardized, open-source benchmarks and datasets is especially evident in the realm of Unmanned Aerial Vehicles (UAVs), where existing evaluation frameworks are largely confined to racing or basic navigation tasks. This limitation not only hinders fair comparison among different cognitive UAV systems but also restricts the exploration of more sophisticated cognitive functions such as reasoning, human recognition, and symbolic understanding.

To address these challenges, we introduce \emph{CognitiveDrone} --- a novel VLA model designed for real-time cognitive task solving and reasoning in UAVs. In conjunction with the model, we propose \emph{CognitiveDroneBench}, an open-source benchmark built upon a Gazebo-based physical simulation environment that integrates a drone racing track with cognitive checkpoints. At each stage of the track, the UAV is required to select specific gates by solving a cognitive task, thus providing a comprehensive performance evaluation that transcends traditional racing metrics.

\begin{figure}[t!]
  \centering
  \includegraphics[width=0.48\textwidth]{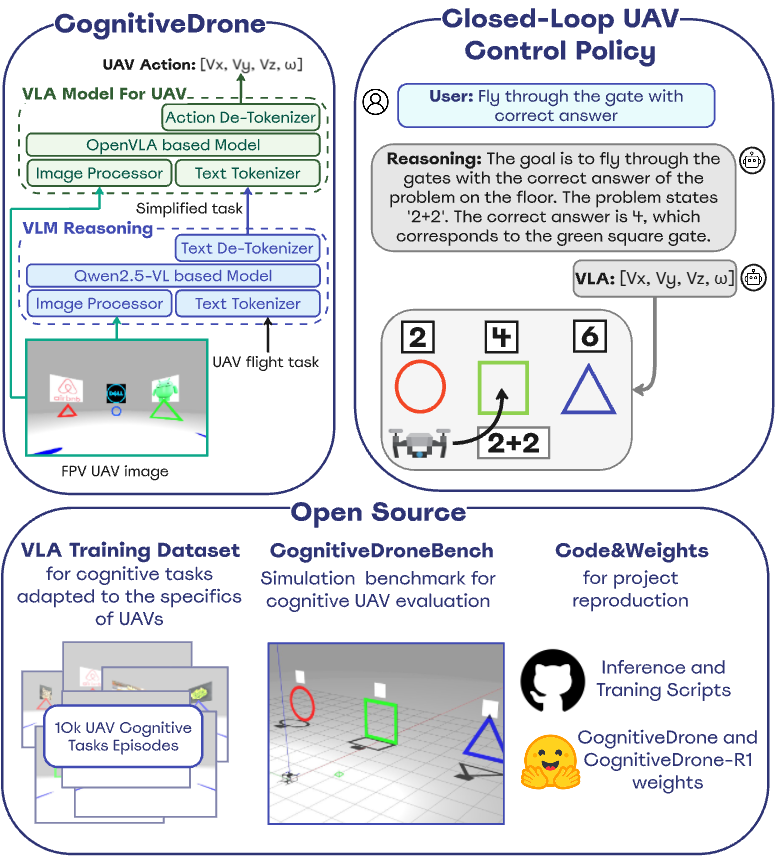}
  \caption{CognitiveDrone is a VLA system for UAVs that generates smooth 4D control commands from first-person visual inputs and natural language instructions. It combines a 7B-parameter VLA model trained on an extensive open-source dataset of cognitive tasks—including reasoning, human recognition, and symbol understanding—with a 7B-parameter VLM reasoning module that refines task directives. The system is evaluated within CognitiveDroneBench—the first evaluation benchmark for VLA systems tailored to cognitive UAVs—where the drone must navigate a track with gates by selecting the appropriate gate through solving cognitive tasks. We have released the complete dataset, benchmark environment, model weights, and training/inference code as open source.}

  \label{fig:overview}
\end{figure}

Furthermore, we augment our system with an auxiliary reasoning module based on the VLM model Qwen2.5-VL, yielding the CognitiveDrone-R1 variant. This reasoning module, operating at a lower frequency than the primary VLA component, is intended to enhance task comprehension and facilitate more robust decision-making. By seamlessly integrating these components, our work paves the way for more rigorous evaluations and innovative applications in cognitive UAV research.

\section{Related Works}\label{sec:related-works}

A substantial body of research has focused on developing cognitive systems for robotics, where the integration of visual perception, language understanding, and action planning is paramount. For robotic manipulators, state-of-the-art models such as PaLM-E~\cite{driess2023palmeembodiedmultimodallanguage}, RT-1~\cite{brohan2023rt1roboticstransformerrealworld}, RT-2~\cite{brohan2023rt2visionlanguageactionmodelstransfer}, and RT-X~\cite{embodimentcollaboration2024openxembodimentroboticlearning} have significantly advanced the field by incorporating VLA frameworks that enable these robots to handle complex manipulation tasks in dynamic environments. Complementary to these efforts, VLA-based systems like OpenVLA~\cite{kim2024openvlaopensourcevisionlanguageactionmodel}, MiniVLA~\cite{belkhale2024minivla}, and Octo~\cite{octo_2023} have been developed for diverse robotic platforms, extending the application spectrum of cognitive robotics.

The evaluation of such systems has been further supported by simulation benchmarks. For instance, LIBERO~\cite{liu2023libero} provides a simulation environment that standardizes the assessment of cognitive performance in robotic manipulators. In parallel, for robots aimed at human-robot collaboration and domestic assistance — including humanoids such as Tesla Optimus, Agility Robotics~\cite{Agility_Robotics}, and FIGURE, as well as quadrupedal platforms like CognitiveDog~\cite{lykov2024cognitivedoglargemultimodalmodel} and DoggyBot~\cite{wu2024helpfuldoggybotopenworldobject} — the open-source PARTNR benchmark~\cite{chang2024partnrbenchmarkplanningreasoning} has been introduced to facilitate objective comparisons.

Within the realm of cognitive UAVs, various transformer-based and VLA approaches have been explored. UAVs Meet LLMs~\cite{tian2025uavsmeetllmsoverviews} presents a comprehensive overview of such methods, highlighting applications in drone navigation~\cite{fan2023aerialvisionanddialognavigation, gao2024aerialvisionandlanguagenavigationsemantictopometric, EmbodiedCityzhang2024, wang2024realisticuavvisionlanguagenavigation, liu2023aerialvlnvisionandlanguagenavigationuavs}, flight control~\cite{vemprala2023chatgptroboticsdesignprinciples, zhong2023safervisionbasedautonomousplanning}, and mission planning~\cite{pan2024vlp, sautenkov2025uavvlavisionlanguageactionlargescale}. Additionally, models for drone swarm control, such as SwarmGPT~\cite{jiao2023swarmgptcombininglargelanguage} and FlockGPT~\cite{lykov2024flockgptguidinguavflocking}, have been developed. SwarmGPT generates individual trajectories for each drone in a swarm, while FlockGPT computes Signed Distance Function (SDF) to define the flight direction for each drone based on its position relative to the target surface. The RaceVLA model~\cite{RaceVLA} notably introduced a four-component control signal $(V_x, V_y, V_z, omega)$ for UAVs, derived from visual inputs and natural language commands. However, the evaluation of RaceVLA has been largely restricted to drone racing tasks, leaving its performance in more complex cognitive scenarios largely unexplored.

In recent times, an alternative direction has emerged that integrates explicit reasoning modules within robotic systems. Initial proposals in language models, such as GPT-4o1 and DeepSeek-R1, as well as models such as Claude Sonnet 3.7, Gemini 2 Thinking, and Grok 3 Big Brain mode, have demonstrated the potential of decoupling reasoning from reactive control. This paradigm has also been adapted in robotics; for example, CognitiveOS~\cite{lykov2024cognitiveoslargemultimodalmodel} implements separate modules for action planning and execution, and the humanoid robot Helix~\cite{figure2025helix} distinguishes between a high-level reasoning component (System 2, powered by a 7B parameter VLM pretrained on extensive internet data) and a low-level fast control component (System 1, employing an 80M parameter visuomotor policy).

Despite these significant contributions, the field of cognitive UAVs still suffers from a lack of open-source datasets and standardized benchmarks capable of evaluating both control and high-level cognitive functions. Our work aims to fill this gap by introducing a novel dataset that encompasses cognitive tasks across various domains (e.g., reasoning, human recognition, symbol understanding) and by providing a comprehensive benchmark — CognitiveDroneBench — tailored to the unique challenges of cognitive UAV platforms.


\section{System Architecture}

\begin{figure}[t!]
\vspace{+3mm}
  \centering
  \includegraphics[width=0.48\textwidth]{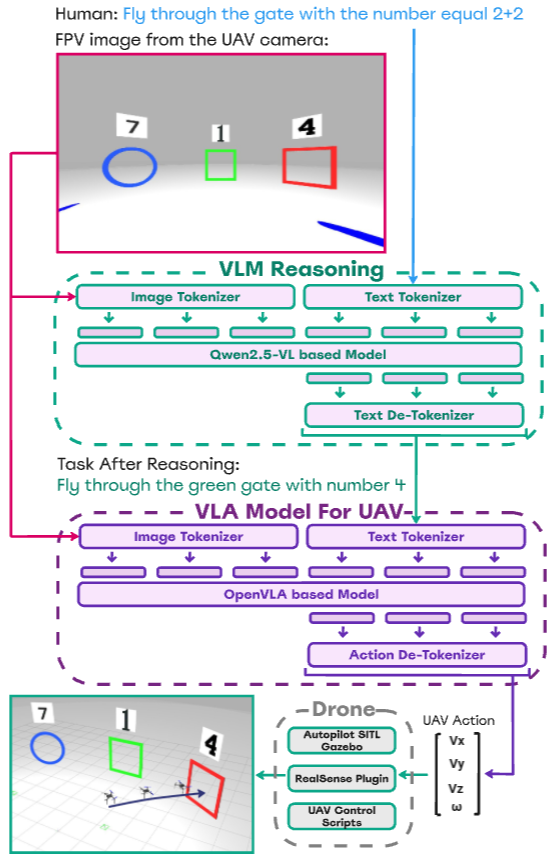}
  \caption{CognitiveDrone system architecture.}
  \label{fig:sys_arch}
\end{figure}

\subsection{CognitiveDrone System Architecture}

Figure~\ref{fig:sys_arch} illustrates the overall architecture of the CognitiveDrone system. The primary goal of our system is to generate smooth 4D action commands for UAVs based on first-person visual inputs and user instructions. To achieve this, we integrate a VLA model adapted from the open-source OpenVLA model, which comprises 7 billion parameters. This model was trained on a dataset of over 8,000 simulated flight episodes, enabling it to develop a robust understanding of UAV flight physics and to effectively control the drone’s motion.

OpenVLA is primarily optimized for accurately capturing the dynamics of drone flight. Consequently, it excels at generating high-frequency control commands—operating at 10 Hz—to produce smooth and continuous trajectories. However, its focus on flight physics means that it may struggle with more complex cognitive tasks, such as resolving ambiguities in task instructions or selecting the most appropriate action when faced with multifaceted challenges.

To address this limitation, our architecture is extended with an additional VLM dedicated to high-level reasoning, forming the CognitiveDrone-R1 system. This reasoning module processes the task instructions and visual inputs to disambiguate and simplify the directives, effectively translating complex commands into clearly defined actions for the VLA model to execute. Owing to the inclusion of a second 7B model for reasoning, the system's total memory requirement doubles to approximately 20 GB, compared to 10 GB when using only the single 7B VLA model. Additionally, due to the higher token count involved in reasoning tasks, the reasoning module operates at a lower frequency—around 2 Hz—while the VLA model continues to generate control commands at 10 Hz.

In summary, our system leverages the complementary strengths of two modules: the high-frequency OpenVLA module, which is specialized in the real-time generation of control commands based on deep flight physics understanding, and the slower, yet powerful, VLM reasoning module, which refines and clarifies task directives. Together, these components enable robust and cognitively capable UAV operation.

\subsection{CognitiveDroneBench Simulation Architecture} 

Gazebo with ArduPilot was chosen for its simulation capabilities, allowing accurate replication of real-world drone dynamics, including flight control; additionally, the drone in simulation is controlled using velocity setpoints, ensuring consistency with real-world drones running ArduPilot firmware.

The simulation environment is built using Gazebo with the ArduPilot Gazebo plugin and SITL ArduPilot.  Each model (gates, labels) for dataset collection is described using an *.sdf file along with a *.dae model. Tasks in the simulation are defined by a JSON file, which includes a task prompt, task options, the correct answer, and gate parameters such as size, shape, and color. The ROS-Gazebo plugin is used to dynamically spawn objects for drone tasks via Python scripts.

During dataset collection, the initial position of the camera drone is spawned with a randomized pose near initial point. A path is then generated from the camera’s initial position to the center of a designated gate, following a spline-based trajectory. As the camera moves along this trajectory, frames with size of 256x256 pixels and applied actions are recorded at each step. The camera simulation is handled using the Gazebo RealSense plugin, with all data collected via ROS topics.

\section{Dataset Collection and Training Pipeline}

\subsection{Dataset Collection}

\begin{figure}[t!]
\vspace{+3mm}
  \centering
  \includegraphics[width=0.48\textwidth]{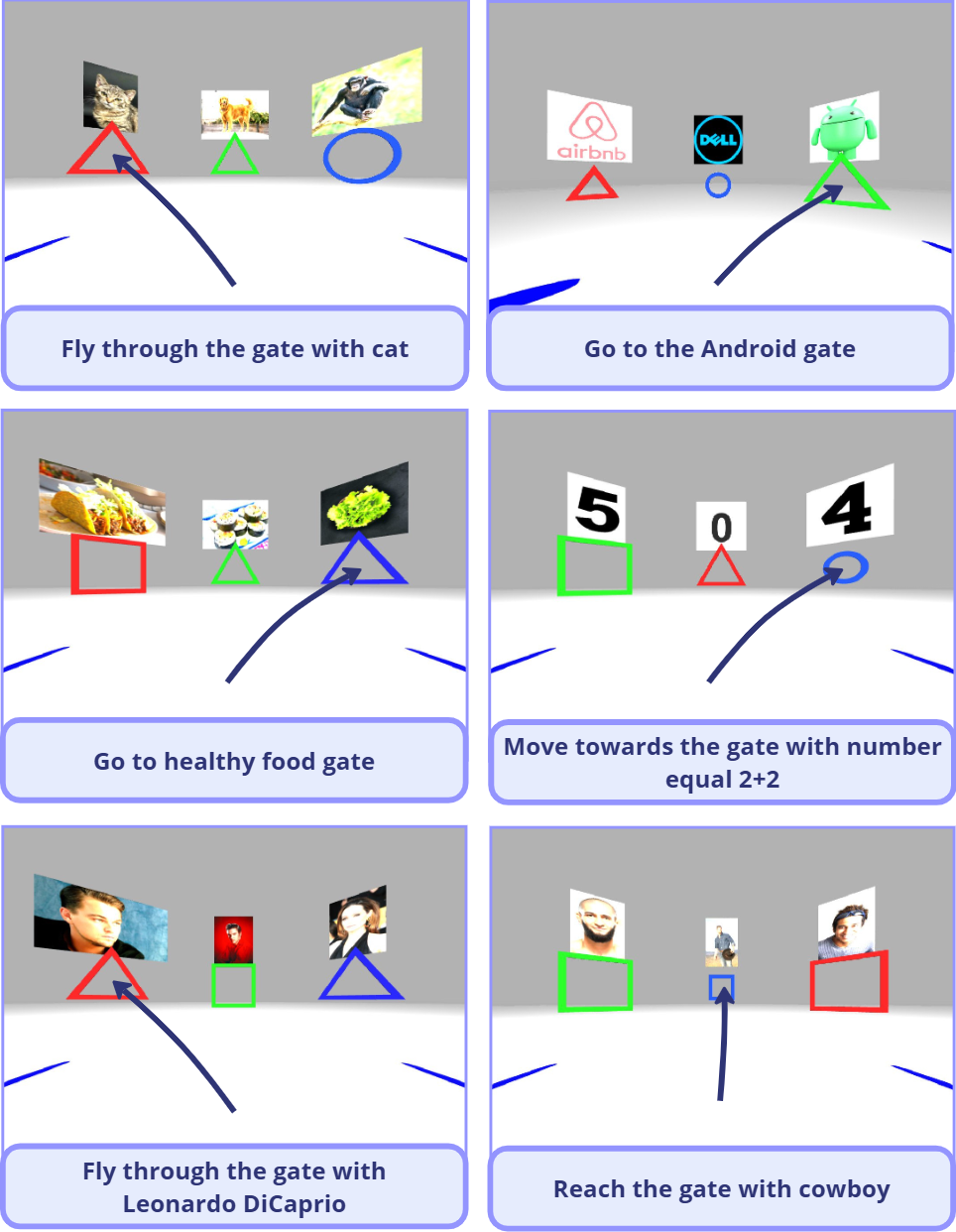}
  \caption{Examples of prepared dataset tasks for VLA to solve cognitive tasks adapted for UAVs.}
  \label{fig:examples}
\end{figure}

We collected a dataset of 8,062 continuous trajectory samples to train the VLA model in a simulated environment, grouped into three categories. Task examples are shown in Figure~\ref{fig:examples}. In all samples, the UAV navigates to a target gate with a preliminary known answer, generating a four-dimensional (4D) action command to control its motion.

\textbf{Human Recognition}: The model is required to identify the individuals based on external characteristics specified within the textual prompt. Additionally, the UAV is tasked with navigating to a designated gate associated with a notable figure.

\textbf{Symbol Understanding}: The model is required to differentiate between a variety of symbols, including alphanumeric characters (e.g., numbers and letters), corporate logos, and pictorial representations of animals.

\textbf{Reasoning}: In this category, the UAV must execute tasks necessitating logical deduction. Examples include navigating to a gate displaying a digit corresponding to the solution of a mathematical problem or associating an object with a specific attribute (e.g., interpreting the instruction “Navigate to the gate with the sweet drink” as selecting a gate marked with a soda logo).

Virtual gates, corresponding to the dataset samples, were instantiated within the simulation. Throughout each flight, the UAV’s velocity and the head profile $(V_x, V_y, V_z, omega)$ was continuously logged, providing a comprehensive record of its dynamic behavior.

The dataset was divided into training and test subsets, ensuring an equitable distribution of samples across all three categories in both sets. The training subset was employed to optimize the VLA model parameters, while the test subset forms the foundation of the CognitiveDroneBench benchmark for performance evaluation.

\begin{figure}[t!]
\vspace{+3mm}
  \centering
  \includegraphics[width=0.48\textwidth]{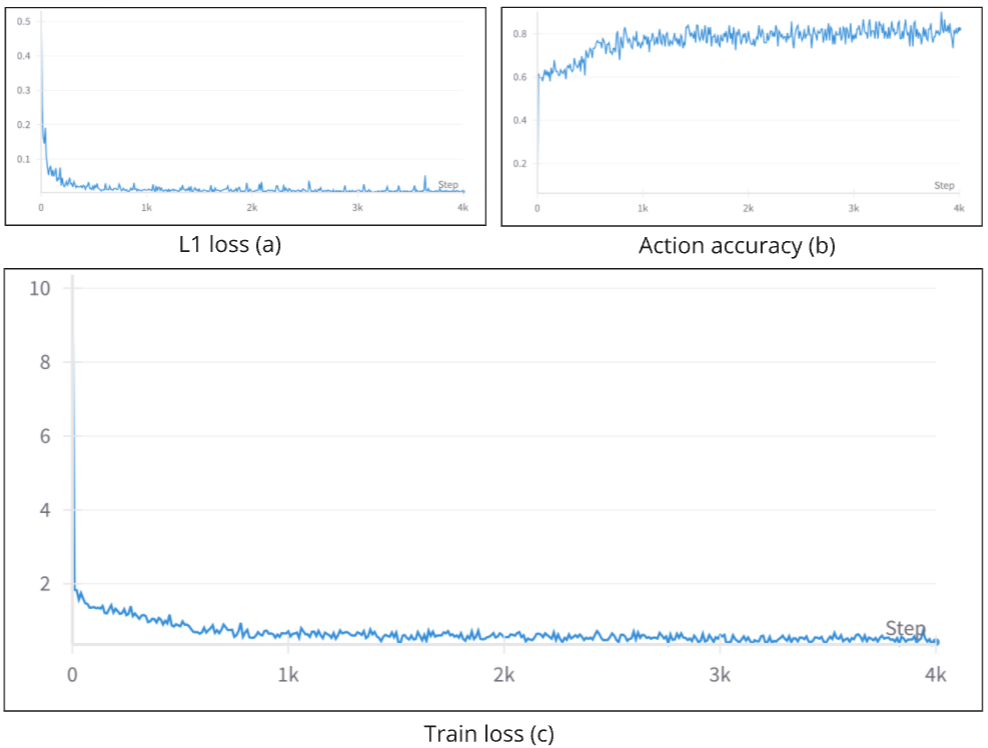}
  \caption{ Metrics Overview: (a) L1 loss indicates absolute prediction errors. (b) Action accuracy quantifies the percentage of correct predictions. (c) Cross-entropy loss measures performance on discretized action tokens.}
  \label{fig:learning_curve}
\end{figure}

\subsection{Model Training}

The VLA model was fine-tuned using our custom training dataset based on the OpenVLA architecture. The collected data was structured in accordance with the Reinforcement Learning Dataset (RLDS) format to ensure seamless compatibility with OpenVLA. This organization facilitated efficient management of actions, images, and task instructions, rendering the dataset suitable for both imitation learning and task-specific action prediction. Subsequently, the dataset was utilized to fine-tune an OpenVLA-7b model employing a parameter-efficient Low-Rank Adaptation (LoRA) approach, applying rank-32 adapters to optimize memory usage while adjusting a minimal set of trainable weights. The training configuration consisted of a batch size of 64, a learning rate of $5 \times 10^{-4}$, and 4000 gradient steps, with image augmentation disabled.

Training was conducted on four NVIDIA A100 GPUs, allowing for efficient large-scale optimization while maintaining real-time deployability on UAV hardware. Checkpoints were saved at regular intervals, with evaluations performed every 500 steps to monitor model performance. For a detailed overview of the training progression, please refer to the learning curve in Figure~\ref{fig:learning_curve}.

\section{Evaluation}

\subsection{CognitiveDroneBench}

The evaluation of cognitive capabilities in UAVs poses unique challenges that are even more pronounced than those encountered in robotic manipulators. To address these challenges and to enable an objective comparison of VLA models for UAVs, we developed an open-source simulation benchmark named \emph{CognitiveDroneBench}. As illustrated in Figure~\ref{fig:overview}, the benchmark is implemented within a high-fidelity physical simulation environment that accurately replicates the dynamics and physics of UAV flight.

In our benchmark, the drone is required to traverse a race track composed of multiple sequential gates. At each stage of the track, the drone receives a first-person view (FPV) image along with a task-specific textual instruction. The core objective is to select the correct gate by solving an embedded cognitive task, thereby generating a 4D action command that dictates its movement. The tasks are categorized into three distinct types: \textbf{Human Recognition}, \textbf{Symbol Understanding}, and \textbf{Reasoning}. 

For conventional VLA models, such as RaceVLA and the base version of CognitiveDrone, the decision-making process involves directly processing the FPV image and associated prompt to determine the appropriate action. In contrast, the CognitiveDrone-R1 variant incorporates an additional reasoning stage using a VLM. This VLM processes the task instruction and visual data to disambiguate and simplify the directive, effectively reducing the complexity of the task before passing it to the VLA model for high-frequency control. 

Task performance is automatically validated: passing through the correct gate earns the drone 1 point. The final score in each category is obtained by normalizing the accumulated points by the maximum achievable score. This methodology provides an objective measure of the cognitive performance of the VLA models under realistic UAV operating conditions.

\begin{figure*}[t!]
\vspace{+3mm}
  \centering
  \includegraphics[width=\textwidth]{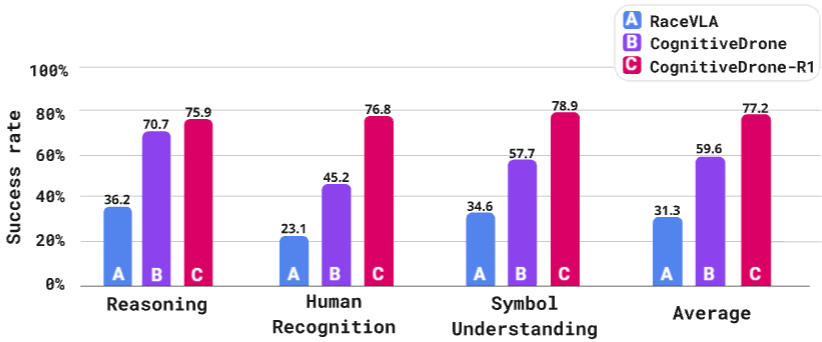}
  \caption{Benchmark performance on CognitiveDroneBench for the RaceVLA, CognitiveDrone, and CognitiveDrone-R1 models. Shown are scores for Reasoning, Human Recognition, and Symbol Understanding tasks, as well as the overall average.}
  \label{fig:benchmark_results}
      \vspace{-4mm}
\end{figure*}

\subsection{Evaluation Methodology and Benchmarking Results}

A comprehensive evaluation was conducted using the CognitiveDroneBench benchmark, with the full results presented in Figure~\ref{fig:benchmark_results}. In this evaluation, the performance of three models—RaceVLA, CognitiveDrone, and CognitiveDrone-R1—was assessed across three cognitive task categories: Human Recognition, Symbol Understanding, and Reasoning. In addition to the individual category scores, an overall "Average" metric was computed as the success rate across all samples in the benchmark.

RaceVLA, optimized primarily for rapid drone racing, demonstrated a strong capability in navigating the physical course due to its robust understanding of UAV flight dynamics. However, its performance in executing cognitive tasks was notably poor, with scores of 36.2\% in Reasoning, 23.1\% in Human Recognition, and 34.6\% in Symbol Understanding, resulting in an overall average success rate of 31.3\%. These results indicate that while RaceVLA reliably guides the drone through the gates, its decision-making when selecting the correct gate based on task instructions is nearly random.

In contrast, the CognitiveDrone model, designed specifically to tackle cognitive challenges, achieved markedly better results: 70.7\% in Reasoning, 45.2\% in Human Recognition, and 57.7\% in Symbol Understanding, yielding an overall average success rate of 59.6\%. This significant improvement, particularly in the Reasoning category, highlights the model’s enhanced ability to process and act upon complex task directives compared to RaceVLA.

Moreover, the integration of a dedicated VLM reasoning module in the CognitiveDrone-R1 system led to substantial performance gains. CognitiveDrone-R1 achieved a success rate of 75.9\% in Reasoning, 76.79\% in Human Recognition, and 78.9\% in Symbol Understanding, with an overall average success rate of 77.2\%. Compared to the base CognitiveDrone model, the reasoning module improved the Reasoning score from 70.7\% to 75.9\% (an increase of about 6\%), raised the Human Recognition score from 45.2\% to 76.8\% (an increase of roughly 31\%), and enhanced the Symbol Understanding score from 57.7\% to 78.9\% (an increase of approximately 21\%). Overall, the addition of the reasoning module resulted in an average success rate gain of around 17.6\%.

In summary, while RaceVLA is effective for basic navigation tasks due to its precise handling of UAV flight physics, it falls short in the nuanced cognitive decision-making required for accurate gate selection. The CognitiveDrone system, particularly when enhanced with an external reasoning module, demonstrates superior performance and robustness across all evaluated categories, underscoring the critical importance of integrating advanced reasoning capabilities into VLA models for UAVs.

\section{Conclusion}

In this work, we introduced the CognitiveDrone VLA model for UAV operations that require advanced cognitive capabilities, along with an enhanced system—CognitiveDrone-R1—that integrates an additional reasoning module to support the drone in solving complex tasks. Experimental evaluations on our open-source benchmark, \emph{CognitiveDroneBench}, demonstrate significant performance improvements. While a racing-oriented model like RaceVLA achieved an overall success rate of only 31.3\%, the CognitiveDrone model increased this figure to 59.6\%. Furthermore, the integration of the reasoning module in CognitiveDrone-R1 elevated the overall success rate to 77.2\%. In particular, CognitiveDrone-R1 improved the reasoning capability by approximately 6\% over the base model, and it boosted Human Recognition and Symbol Understanding performance by roughly 31\% and 21\%, respectively.

These results underscore the critical importance of incorporating advanced reasoning capabilities into VLA models for UAVs, as doing so not only enhances performance in tasks requiring logical inference but also improves overall decision-making across diverse cognitive challenges. Moreover, the introduction of CognitiveDroneBench provides the first dedicated, open-source platform for objectively assessing the cognitive capabilities of VLA models in UAV applications. The complete codebase, dataset, and benchmark are publicly available for further research and development.

\balance
\bibliographystyle{IEEEtran}
\bibliography{ref}

\end{document}